\documentclass[10pt,twocolumn,letterpaper]{article}

\usepackage{cvpr}      
\usepackage{makecell}
\usepackage{threeparttable}
\usepackage{multirow}
\usepackage{algorithmic}
\usepackage[ruled,linesnumbered]{algorithm2e}
\usepackage{indentfirst}
\usepackage[dvipsnames]{xcolor}

\definecolor{cvprblue}{rgb}{0.21,0.49,0.74}
\usepackage[pagebackref,breaklinks,colorlinks,citecolor=cvprblue]{hyperref}

\def\paperID{17406}
\def\confName{CVPR}
\def\confYear{2024}

\title{Boosting the Transferability of Adversarial Examples via Local Mixup and Adaptive Step Size}

\author{Junlin Liu and Xinchen Lyu\\
Beijing University of Posts and Telecommunications\\
{\tt\small zero10walker@gmail.com and lvxinchen@bupt.edu.cn}
}

\begin{document}
\maketitle
\begin{abstract}
    Adversarial examples are one critical security threat to various visual applications, where injected human-imperceptible perturbations can confuse the output.
    Generating transferable adversarial examples in the black-box setting is crucial but challenging in practice.
    Existing input-diversity-based methods adopt different image transformations, but may be inefficient due to insufficient input diversity and an identical perturbation step size.
    Motivated by the fact that different image regions have distinctive weights in classification, this paper proposes a black-box adversarial generative framework by 
    jointly designing enhanced input diversity and adaptive step sizes.
    We design local mixup to randomly mix a group of transformed adversarial images, strengthening the input diversity.
    For precise adversarial generation, we project the perturbation into the $tanh$ space to relax the boundary constraint. 
    Moreover, the step sizes of different regions can be dynamically adjusted by integrating a second-order momentum.
    Extensive experiments on ImageNet validate that our framework can achieve superior transferability compared to state-of-the-art baselines.
\end{abstract}    
\section{Introduction}
\label{sec:intro}

Deep learning models have demonstrated tremendous performance in various visual tasks, including image classification~\cite{he2016identity,simonyan2015very,szegedy2016rethinking,xie2017aggregated}, face recognition~\cite{wang2018cosface,song2022adaptive}, and self-driving~\cite{sallab2017deep,mori2019visual}.
However, the trained models are vulnerable to adversarial examples~\cite{szegedy2013intriguing}, which are crafted carefully by injecting human-imperceptible perturbations.
Adversarial examples can make the deployed models deviate from the normal prediction and may cause serious security threats~\cite{eykholt2018robust,wang2022triangle,yuan2022natural}, \eg, recognizing the stop sign as the speed limit in self-driving.

Based on the adversary's knowledge of the victim model, the adversarial examples can be categorized into white-box~\cite{goodfellow2015shlens,moosavi2016deepfool} or black-box attacks~\cite{zhang2021beyond,xiong2022stochastic,chen2023adaptive}. 
The white-box attacks assume that the adversary has the full knowledge of the target model to generate the adversarial examples.
In contrast, the black-box attacks consider a more realistic setting, where the adversary cannot get access to the target model directly.
Black-box attacks are of practical importance and have attracted broad attention.

There are two types of black-box adversarial attacks, \ie, \textit{query-based}~\cite{suya2020hybrid,wang2022triangle} and \textit{transfer-based}~\cite{long2022frequency,xiong2022stochastic,chen2023adaptive,wang2023structure} attacks.
1) Query-based adversarial attacks send queries to the target model and collect information to craft adversarial examples, which may be detected by the service provider.
2) Transfer-based attacks generate adversarial examples from the white-box surrogate models by exploiting the cross-model transferability of adversarial examples.
Transfer-based attacks assume no a-priori knowledge of the target model, and are promising and stealthy to real-world AI applications. 

The critical challenge of transfer-based attacks is to improve the adversarial transferability  (\ie adversarial success rate crafted from surrogate models).
Various methods have been designed in the literature, including input diversity~\cite{xie2019improving,zou2020improving,Dong_2019_CVPR,wu2021improving,wang2023structure}, optimization improvement~\cite{dong2018boosting,lin2019nesterov,wang2021enhancing},  ensemble attacks~\cite{xiong2022stochastic,chen2023adaptive} and advanced loss design~\cite{li2020towards,zhao2021success}.
Input diversity achieves superior transferability by incorporating image transformations into the synthesis process, hence alleviating overfitting to surrogate models.
Existing methods mainly focus on increasing diversity by adding global~\cite{zou2020improving,wu2021improving} or local~\cite{wang2023structure} transformation categories.

However, existing input-diversity-based works~\cite{xie2019improving,zou2020improving,Dong_2019_CVPR,wu2021improving,wang2023structure} adopt an identical step size to craft adversarial examples. 
We find that 1) input diversity can be strengthened by mixing the transformed images; 
2) identical step size may degrade the adversarial transferability (as different regions of an image are of different weights for model inference).
Motivated by this, we jointly design enhanced input diversity and adaptive step size to boost the transferability of black-box adversarial examples.

This paper proposes a novel adversarial generative framework, \underline{I}nput-\underline{D}iversity-based \underline{A}daptive \underline{A}ttack (IDAA), by jointly designing enhanced input diversity and adaptive step size.
\textit{1) Enhanced input diversity.} 
We exploit different image transformation techniques to reformulate a group of adversarial examples to obtain input diversity.
To further increase the input diversity, we design a \textit{local mixup} module (\ie randomly mixing the regions of the group of transformed images) before feeding to the surrogate model.
\textit{2) Adaptive step size.}
We propose to project the perturbation optimization into the $tanh$ space to relax the boundary constraint. 
The step sizes can be dynamically adjusted by applying a second-order momentum.

Our main contributions can be summarized as follows.
\begin{itemize}
    \item We propose to design the local mixup step (that randomly mixes a group of image regions) to strengthen the input diversity.
    \item For precise adversarial generation, we eliminate the constraint for image validity and perturbation budget by projecting into the $tanh$ space. 
    We also enable adaptive adjustment of step sizes for different perturbation regions during the update phase via second-order momentum.
    \item By integrating the proposed local mixup and adaptive step size, we design an adversarial generative framework, IDAA, to craft highly transferable adversarial examples.
    \item Experimental results on ImageNet validate that our proposed framework can achieve superior performance in comparison to stat-of-the-art baselines. 
    Our framework can also work in conjunction with other transferable methods to further improve their transferability.
\end{itemize}
\section{Related Works}
\label{sec:relate}

\begin{figure*}[th]
    \centering
    \includegraphics[width=0.8\linewidth, height=0.4\linewidth]{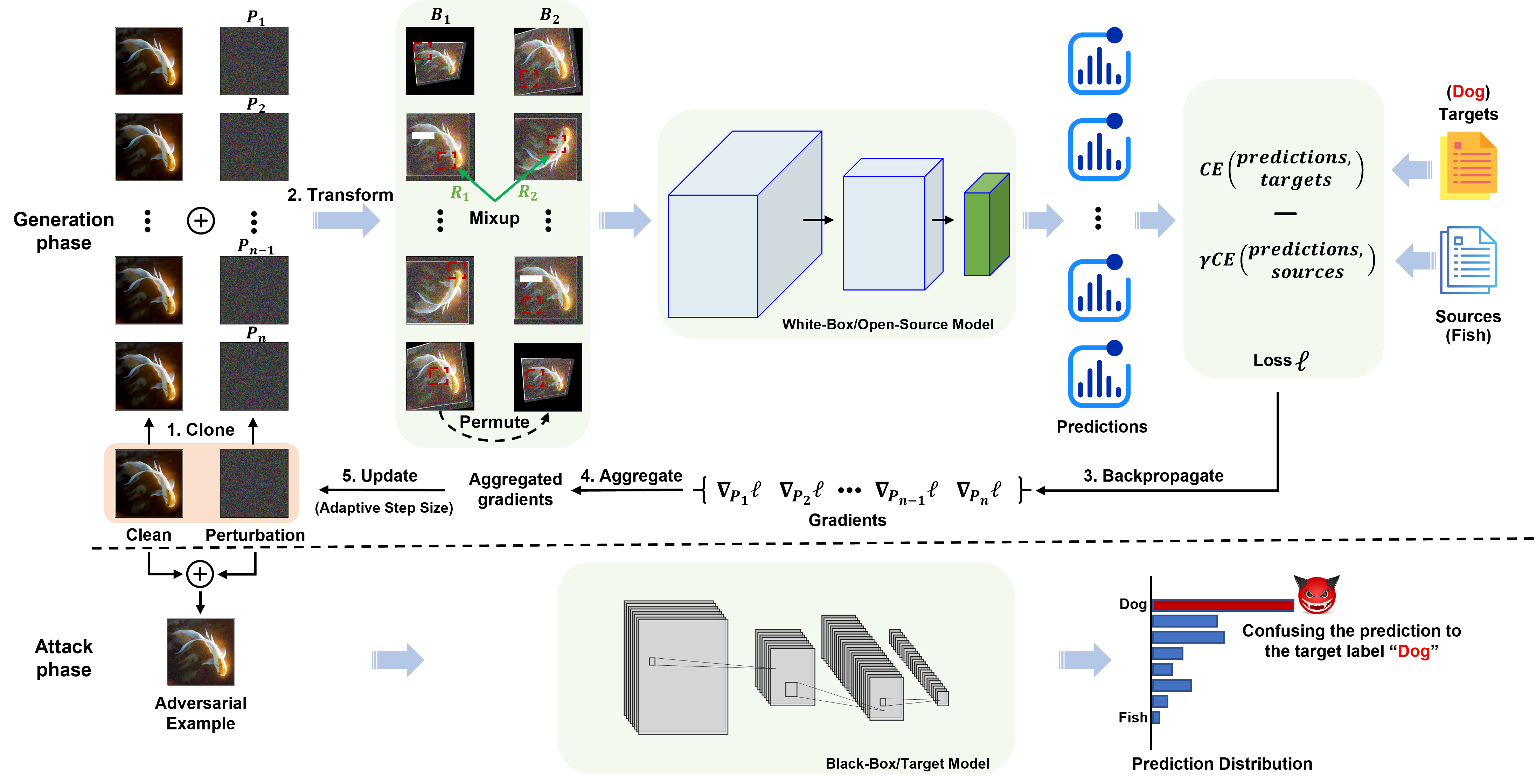}
    \vskip -0.1in
    \caption{Overview of IDAA. Before sending to the surrogate model, the crafted adversarial examples are transformed by various image transformations. Then, the random regions of the transformed variants are mixed to further strengthen the input diversity.}
    \label{fig:overview}
    \vskip -0.15in
\end{figure*}

Adversarial attacks attempt to manipulate AI models to deviate from the normal output by carefully crafted input data (\ie, adversarial examples).
Based on the deviation direction, adversarial attacks can be categorized into \textit{untargeted}~\cite{moosavi2016deepfool,wu2019untargeted} and \textit{targeted}~\cite{inkawhich2019feature,li2020towards,zhao2021success} adversarial attacks.
Targeted adversarial attacks attempt to obtain some specific model output, while untargeted adversarial attacks aim to confuse the model to make mistakes. 

In this paper, we focus on transfer-based black-box adversarial attacks, \ie, the adversary exploits the cross-model transferability to conduct attacks. 
Specifically, transfer-based attacks exploit a white-box/open-source surrogate model to synthesize adversarial examples and leverage the crafted adversarial examples to confuse the unknown target models via the transferability of adversarial examples. 
The transferability of adversarial examples is critical to achieving effective attack performance for transfer-based adversarial attacks.
Several methods have been proposed to boost adversarial transferability.

\textbf{Optimization improvement}.
Dong et al.~\cite{dong2018boosting} proposed to integrate the momentum term into the iterative update, which can help to stabilize the update direction and escape from getting into the local maxima. 
Lin et al.~\cite{lin2019nesterov} exploited the looking ahead property of Nesterov Accelerated Gradient to improve the adversarial transferability.
Instead of only considering the current gradient, Xiaosen Wang and Kun He~\cite{wang2021enhancing} proposed to tune the gradient through the gradient variance.

\textbf{Input diversity}.
The generated adversarial examples may overfit the white-box surrogate model easily, degrading the adversarial transferability. 
To alleviate the overfitting, Xie et al.~\cite{xie2019improving} exploited random resizing to transform the input image with probability $p$ in each iteration.
Different from  Xie et al.~\cite{xie2019improving}, Zou et al.~\cite{zou2020improving} set the transformation probability $p=1$, a larger resizing scale and finally resized the transformed images back to the original size, which further boosts the transferability.
Dong et al.~\cite{Dong_2019_CVPR} shifted the image pixels to translate the input image and exploited the translated images to generate transferable adversarial examples, which are less sensitive to the discriminative regions of the surrogate model.
To further increase diversity, Wu et al.~\cite{wu2021improving} proposed to exploit an adversarial transformation network to transform the input image.
Wang et al.~\cite{wang2023structure} showed that more transferability can be obtained by applying different transformations to different blocks of an image randomly. 
Long et al.~\cite{long2022frequency} explored the image transformation in the frequency domain by Gaussian noise and random mask.

\textbf{Ensemble attacks}.
Besides input diversity, the adversary can also exploit multiple models to generate the adversarial examples~\cite{dong2018boosting}.
Reducing the gradient variance of the ensemble models benefits improving the adversarial transferability~\cite{xiong2022stochastic}.
To craft the adversarial examples, the outputs of different models in ensemble attacks usually need to be aggregated.  
Chen et al.~\cite{chen2023adaptive} demonstrated that the aggregation should consider the contribution to the adversarial objects to boost the transferability. 

\textbf{Advanced loss design}.
Some other loss functions may be more suitable for generating transferable adversarial examples in comparison to the cross-entropy loss.
Li et al.~\cite{li2020towards} exploited Poincar{\'e} distance as loss criteria to make the magnitude of the gradient increase with getting closer to the target label, which makes the gradient self-adaptive.
Besides, they also combined an angular-distance-based triplet loss to further strengthen the transferability.  
Zhao et al.~\cite{zhao2021success} found that the logit loss can obtain stronger gradients in comparison to the cross-entropy loss, also improving the adversarial transferability. 

\section{Methodology}
In this section, we elaborate on the proposed adversarial generative framework, IDAA, improving the adversarial transferability by enhanced input diversity and adaptive step size.
The attack pipeline of IDAA is shown in \cref{fig:overview}.
We first detail the attack objective, and then introduce the three key steps for boosting adversarial transferability, \ie, local mixup, tanh projection and adaptive step size.

\subsection{Attack Objective}
Adversarial examples $x_{adv}$ aim to confuse the target model $f_b$ to deviate from the ground-truth label $y_{src}$ or output the target label $y_{tgt}$. 
Attackers craft adversarial examples by injecting human-imperceptible perturbations $r$ into the benign inputs $x$, i.e., $x_{adv} = x + r$. 
Typically, $l_{\infty}$-norm is adopted to constrain the adversarial perturbations $r$ with budget $\epsilon$, i.e., $\Vert r \Vert_{\infty} \leq \epsilon$.
With a surrogate model $f_w$, attackers optimize the adversarial perturbation as 
\begin{equation}
    r_{t+1} = r_t - \alpha\nabla_{r_t}\mathcal{L}(f_w(x+r_t), y_{src}, y_{tgt})
    \label{eq:adv-opt}
\end{equation}
where $\mathcal{L}$, $r_t$ and $\alpha$ denote the loss function, perturbation in step $t$, and step size, respectively.
We focus on generating targeted adversarial examples, which can be measured by fooling success rate (\ie, $f_b(x_{adv}) \neq y_{src}$) and targeted success rate (\ie, $f_b(x_{adv}) = y_{tgt}$) simultaneously. 
The attacker attempts to manipulate the victim model to output the target label. 
The loss function is
\begin{equation}
    \mathcal{L}(f_w(x+r_t), y_{src}, y_{tgt}) = \text{loss}(f_w(x+r_t), y_{tgt})
    \label{eq:targeted}
\end{equation}
where $loss$ is the cross-entropy loss.
To avoid the generated adversarial examples too close to the original class~\cite{li2020towards}, we exploit a triplet loss to conduct adversarial attacks, which is expressed as 
\begin{equation}
    \begin{split}
        & L_{pos} = \text{loss}(f_w(x+r_t), y_{tgt}) \\
        & L_{neg} = \text{loss}(f_w(x+r_t), y_{src}) \\
        & \mathcal{L}(f_w(x+r_t), y_{tr}, y_{tgt}) = L_{pos} - \gamma L_{neg} 
    \label{eq:triplet-targeted}
    \end{split}
\end{equation}
where $\gamma$ is the weight controlling deviation strength.

\subsection{Transformation and Local Mixup}
\begin{figure*}
    \centering
    \includegraphics[width=\linewidth]{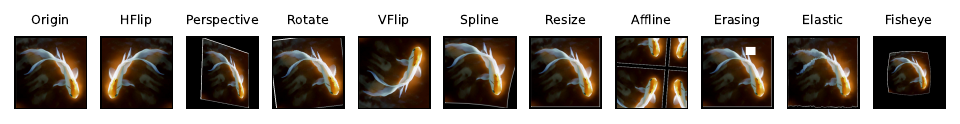}
    \vskip -0.1in
    \caption{Various differential image transformations adopted by IDAA. \textit{Origin} is the raw image before the transformation.}
    \label{fig:trans}
    \vskip -0.15in
  \end{figure*}

Before feeding the crafted inputs to the surrogate model, we first apply various transformations to pre-process them, as diverse input patterns improve the transferability.
The adopted image transformations are shown in \cref{fig:trans}.
To further strengthen the input diversity, we design a \textit{local mixup} step, which randomly selects two patches of the same size from two pre-processed adversarial examples and mixes them~\cite{zhang2018mixup}.

In particular, after conducting the various image transformations, several transformed variants $\boldsymbol{B}_1$ of the adversarial example are generated.
We randomly permute these generated variants to obtain another group of transformed adversarial examples $\boldsymbol{B}_2$. 
Then, we align these two groups of adversarial examples based on the indexes.
For the first pair of adversarial examples $\boldsymbol{B}_1^1$ and $\boldsymbol{B}_2^1$, we randomly sample two local regions $\boldsymbol{R}_1$ and $\boldsymbol{R}_2$ from $\boldsymbol{B}_1^1$ and $\boldsymbol{B}_2^1$, respectively.
We mix these two regions and insert back to the region $\boldsymbol{R}_1$.
The mix operation is conducted as
\begin{equation}
    \boldsymbol{R}_1^{\prime} = \lambda \cdot \boldsymbol{R}_1 + (1 - \lambda) \cdot \boldsymbol{R}_2
\end{equation}
where $\lambda$ is sampled from the beta distribution.
The transformed examples are mixed in a pairwise manner.
After the mixup operation, the mixed $\boldsymbol{B}_1^{\prime}$ is fed to the surrogate model.
Thus, \cref{eq:adv-opt} can be converted to 
\begin{equation}
    r_{t+1} = r_t - \frac{\alpha}{|\boldsymbol{B}_1^{\prime}|} \sum_{x^{\prime} \in \boldsymbol{B}_1^{\prime}} \nabla_{r_t} \mathcal{L}(f_w(x^{\prime}), y_{src}, y_{tgt})
    \label{eq:adv-opt-mix}
\end{equation}

\subsection{Tanh Projection}
To ensure the crafted adversarial images are valid and meet the perturbation budget $\epsilon$, the injected perturbation $r$ should satisfy $0 \leq x + r \leq 1$ and $\Vert r \Vert_{\infty} \leq \epsilon$.
Thus, two clip operations are usually adopted to clip the adversarial examples, \ie, $\text{clip}_0^1(x+\text{clip}_{-\epsilon}^{\epsilon}(r))$, where $\text{clip}_a^b$ denotes the result is clipped within the range from $a$ to $b$. 
The clip operations may cause a negative influence on the perturbation optimization (\eg, getting stuck in a flat spot)~\cite{carlini2017towards}.

To avoid the clip operations, we will not directly optimize the perturbation.
Given the input image $x$ and the perturbation budget $\epsilon$, we can compute the lower and upper perturbation boundary ahead.
Then, we propose to eliminate the clip operation by \textit{tanh} transformation.
Specifically, we first obtain the perturbation boundary as
\begin{equation}
\begin{split}
    Bnd_{lower} &= -\text{min}(x, \epsilon) \\
    Bnd_{upper} &= \text{min}(1 - x, \epsilon) 
    \label{eq:pert-boundary}
\end{split}
\end{equation}
After obtaining the lower and upper boundary, we can express the perturbation $r$ as
\begin{equation}
    r = Bnd_{lower} + Bnd \cdot (tanh(w) / 2.0 + 0.5)
    \label{eq:new-pert}
\end{equation}
where $Bnd = Bnd_{upper} - Bnd_{lower}$ and $-\infty \textless w \textless \infty$.
Through the above transformation, the generated perturbation will satisfy the constraints $0 \leq x + r \leq 1$ and $\Vert r \Vert_{\infty} \leq \epsilon$. 
Thus, instead of optimizing the perturbation $r$ directly, we can choose to update $w$ in the tanh space via unrestrained optimization efficiently.
\cref{eq:adv-opt-mix} in the $tanh$ space (\ie, $w$) can be reformulated as 
\begin{equation}
    w_{t+1} = w_t - \frac{\alpha}{|\boldsymbol{B}_1^{\prime}|} \sum_{x^{\prime} \in \boldsymbol{B}_1^{\prime}} \nabla_{w_t} \mathcal{L}(f_w(x^{\prime}), y_{src}, y_{tgt})
    \label{eq:new-adv-opt-mix}
\end{equation}
After the optimization step, we can obtain $w^{*}$ and compute the perturbation through \cref{eq:new-pert}. 
The generated adversarial example is crafted as
\begin{equation}
    x_{adv} = x + (Bnd_{lower} + Bnd \cdot (tanh(w^{*}) / 2.0 + 0.5))
    \label{eq:new-adv}
\end{equation}

\subsection{Adaptive Step Size}
Existing adversarial attacks~\cite{xie2019improving,zou2020improving,Dong_2019_CVPR,wu2021improving,wang2023structure} adopt an identical step size to update the perturbation. 
This may result in transferability degradation as different image regions may have distinct impacts/weights on the model classification output. 
To this end, we integrate the second-order momentum to adaptively adjust the step size besides the first-order momentum.

The average gradient in \cref{eq:new-adv-opt-mix} can be written as
\begin{equation}
    g_t = \frac{1}{|\boldsymbol{B}_1^{\prime}|}\sum_{x^{\prime} \in \boldsymbol{B}_1^{\prime}} \frac{\nabla_{w_t}\mathcal{L}(f_w(x^{\prime}), y_{src}, y_{tgt})}{\Vert \nabla_{w_t}\mathcal{L}(f_w(x^{\prime}), y_{src}, y_{tgt}) \Vert_1}
    \label{eq:gd}
\end{equation}
The first-order and second-order momentums~\cite{KingBa15} are
\begin{equation}
    \begin{split}
        m &= \beta_1 m + (1 - \beta_1) g_t\\
        v &= \beta_2 v + (1 - \beta_2) g_t^2  
    \end{split}
    \label{eq:momentum}
\end{equation}
where $\beta_1$ and $\beta_2$ are the exponential decay coefficients.
The second-order momentum $v$ is exploited to adjust the step size $\alpha$ dynamically.
Finally, \cref{eq:new-adv-opt-mix} with the first/second-order momentums become to 
\begin{equation}
    w_{t+1} = w_t - \frac{\alpha}{\sqrt{v}}m 
    \label{eq:new-adp-adv-opt-mix}
\end{equation}

\cref{alg:idaa-attack} summarizes the proposed attack framework.
\setlength{\textfloatsep}{0.2in}
\begin{algorithm}[tb]
	\caption{IDAA Attack Algorithm.}
	\label{alg:idaa-attack}
	\KwIn{clean image $x$, ground-truth label $y_{src}$, target label $y_{tgt}$, steps $T$, surrogate model $f_w$, perturbation budget $\epsilon$, group size $N$, and loss function $\mathcal{L}$.}
	\KwOut{adversarial example $x_{adv}$}
	
    computing the perturbation boundary by \cref{eq:pert-boundary}\;
    initialing $w_1 \sim Gaussian(0, 1)$\;
    cloning $N$ clean images $x^N$\;
    $m=0, v=0$\;
	\For{t = 1 $\rightarrow$ $T$}{
        cloning $N$ perturbation variable $w_t^N$\; 
        computing $N$ perturbations $r_t^N$ by \cref{eq:new-pert}\;
        synthesizing $N$ adversarial images $x_{adv}^N = x^N + r_t^N$\;
        circularly transforming the images by the transformations in \cref{fig:trans}\;
        randomly mixing the regions of the transformed image set $B_1^{\prime}$\; 
        calculating the gradient $g_t$ by \cref{eq:gd}\;
        computing $m$ and $v$ by \cref{eq:momentum}\;
        computing $w_{t+1}$ by \cref{eq:new-adp-adv-opt-mix}
	}
    crafting the adversarial example $x_{adv}$ by \cref{eq:new-adv}\;
    \KwRet{$x_{adv}$}
\end{algorithm}

\begin{table*}[tbp]
    \begin{center}
    \small 
            \begin{tabular}{lccccccccccc}
            \toprule \toprule
            \multirow{4}{*}{Surrogate} & \multirow{4}{*}{Attack} & \multicolumn{10}{c}{Target Model} \\
            \cmidrule(lr){3-12}
            & & \multicolumn{2}{c}{RN-50} & \multicolumn{2}{c}{DN-121} & \multicolumn{2}{c}{WRN50-2} & \multicolumn{2}{c}{VGG-19} & \multicolumn{2}{c}{Inc-v3} \\
            \cmidrule(lr){3-4}
            \cmidrule(lr){5-6}
            \cmidrule(lr){7-8}
            \cmidrule(lr){9-10}
            \cmidrule(lr){11-12}
            & & fSuc & tSuc & fSuc & tSuc & fSuc & tSuc & fSuc & tSuc & fSuc & tSuc\\
            \midrule \midrule
            \multirow{6}{*}{RN-50} & DIM & 98.50* & 93.39* & 57.62 & 9.72 & 50.40 & 9.02 & 61.72 & 3.91 & 31.26 & 1.90 \\ 
            & Po-TRIP & 99.20* & 93.09* & 53.81 & 10.62 & 47.9 & 8.92 & 61.72 & 4.51 & 30.56 & 2.10 \\ 
            & MI & 98.90* & 97.80* & 30.66 & 1.20 & 28.76 & 1.50 & 42.89 & 0.40 & 16.23 & 0.10 \\ 
            & TIM & 98.90* & 93.89* & 60.62 & 11.02 &	58.62 & 12.93 &	69.44 & 4.41 & 36.97 & 2.20 \\ 
            & SIT & \textbf{99.90*} & 99.30* & 76.55 & 34.87 & 73.05 & 37.98 & 75.45 & 14.73 & 41.48 &6.41\\ 
            & \textbf{IDAA} & \textbf{99.90*} & \textbf{99.50*} & \textbf{90.78} & \textbf{42.48} & \textbf{88.08} & \textbf{46.29} & \textbf{88.98} & \textbf{18.94} & \textbf{64.63} & \textbf{14.63}\\
            \hline \hline
            \multirow{6}{*}{DN-121} & DIM & 51.10 & 6.01 & 99.60* & 98.70* & 44.19 & 4.61	& 63.43 & 3.51 & 29.06 & 1.70 \\
            & Po-TRIP & 47.90 & 6.51 & 99.70* & 98.60* & 41.88 & 5.51 & 58.82 & 3.01 & 28.36 & 2.40 \\ 
            & MI & 35.27 & 1.40 & 99.80* & 99.70* & 29.86 & 1.20 & 47.19 & 1.10 & 15.53 & 0.30\\ 
            & TIM & 54.01 & 6.51 & 99.80* & 99.00* & 49.00 & 5.41 & 67.54 & 2.51 & 32.36	& 1.60\\ 
            & SIT & 73.85 & 31.06 & \textbf{100.00*} & \textbf{100.00*} & 64.23 & 23.45 & 76.75 & 14.23 & 41.58 & 7.31 \\ 
            & \textbf{IDAA} &  \textbf{88.68} & \textbf{41.28} & \textbf{100.00*} & 99.90* & \textbf{83.77} & \textbf{33.27} & \textbf{89.28} & \textbf{19.84} & \textbf{69.84} & \textbf{18.54}\\
            \hline \hline
            \multirow{6}{*}{WRN50-v2} & DIM & 58.12 & 7.41 & 56.11 & 6.41 & 95.79* & 85.67* & 61.22 & 2.61 & 30.66 & 1.50\\
            & Po-TRIP & 55.81 & 9.32 & 54.01 & 7.52 & 97.90* & 88.18* & 56.91 & 3.61 & 30.16	& 1.90\\
            & MI & 37.78 & 1.80 & 33.77 & 1.00 & 93.29* & 88.98* & 44.19 & 0.40 & 17.03 &0.10\\ 
            & TIM & 60.82 & 7.62 & 56.61 & 7.01 & 97.80* & 89.58* & 65.33 & 3.01 & 36.57 & 1.70\\ 
            & SIT & 79.76 & 36.37 & 76.35 & 24.25 & 99.90* & 96.79* & 73.55 & 9.72 & 43.99 & 5.01\\ 
            & \textbf{IDAA} & \textbf{91.58} & \textbf{43.89} & \textbf{88.78} & \textbf{33.67} & \textbf{100.00*} & \textbf{98.70*} & \textbf{86.87} & \textbf{14.03} & \textbf{67.84} & \textbf{14.53}\\ 
            \hline \hline
            \multirow{6}{*}{VGG-19} & DIM & 38.78 & 1.00 & 39.68 & 1.00 & 31.06 & 0.60 & 99.40* & 94.49* & 21.04 & 0.30\\ 
            & Po-TRIP & 37.47 & 1.40 & 37.68 & 2.00 & 27.56 & 0.20 & 99.70* & 94.59* & 19.64 & 0.20\\ 
            & MI & 28.26 & 0.20 & 28.26 & 0.10 & 22.55 & 0.10 & 98.80* & 94.29* & 15.33 & 0.00\\ 
            & TIM & 46.79 & 1.40 & 45.79 & 1.50 & 37.78 & 0.70 & 99.20* & 96.69* & 27.96 & 0.20\\ 
            & SIT & 49.40 & 4.31 & 51.30 & 6.21 & 38.68 & 3.01 & 99.90* & 97.70* & 26.85 & 1.80\\ 
            & \textbf{IDAA} & \textbf{68.54} & \textbf{12.73} & \textbf{72.85} & \textbf{13.73} & \textbf{57.11} & \textbf{7.52} & \textbf{100.00*} & \textbf{99.00*} & \textbf{45.59} & \textbf{5.01}\\ 
            \bottomrule \bottomrule
            \end{tabular}
    \caption{Fooling and targeted success rates (\%) of the adversarial examples generated by RN-50, DN-121, WRN50-2 and VGG-19. ``*'' indicates the white-box setting.}
    \label{tab:nat-attack}
    \end{center}
    \vskip -0.2in
    \end{table*}
\section{Experiments}
    \begin{table*}[tbp]
        \begin{center}
        \small
                \begin{tabular}{lccccccccccc}
                \toprule \toprule
                \multirow{4}{*}{Surrogate} & \multirow{4}{*}{Attack} & \multicolumn{10}{c}{Target Model} \\
                \cmidrule(lr){3-12}
                & & \multicolumn{2}{c}{RN-50} & \multicolumn{2}{c}{DN-121} & \multicolumn{2}{c}{WRN50-2} & \multicolumn{2}{c}{VGG-19} & \multicolumn{2}{c}{Inc-v3} \\
                \cmidrule(lr){3-4}
                \cmidrule(lr){5-6}
                \cmidrule(lr){7-8}
                \cmidrule(lr){9-10}
                \cmidrule(lr){11-12}
                & & fSuc & tSuc & fSuc & tSuc & fSuc & tSuc & fSuc & tSuc & fSuc & tSuc\\
                \midrule \midrule
                \multirow{6}{*}{RN-50} & DIM & 98.50* & 93.39* & 57.62 & 9.72 & 50.40 & 9.02 & 61.72 & 3.91 & 31.26 & 1.90 \\ 
                & \textbf{DA-DIM} & \textbf{100.00*} & \textbf{100.00*} & \textbf{88.58} & \textbf{31.86} & \textbf{86.47} & \textbf{34.47} & \textbf{86.17} & \textbf{12.22} & \textbf{62.53} & \textbf{8.52}\\ 
                & TIM & 98.90* & 93.89* & 60.62 & 11.02 &	58.62 & 12.93 &	69.44 & 4.41 & 36.97 & 2.20 \\ 
                & \textbf{DA-TIM} & \textbf{100.00*} & \textbf{100.00*} & \textbf{90.68} & \textbf{39.28} & \textbf{89.68} & \textbf{45.59} & \textbf{89.28} & \textbf{14.33} & \textbf{66.93} & \textbf{11.52}\\ 
                & SIT & 99.90* & 99.30* & 76.55 & 34.87 & 73.05 & 37.98 & 75.45 & 14.73 & 41.48 &6.41\\ 
                & \textbf{DA-SIT} & \textbf{99.90*} & \textbf{99.40*} & \textbf{89.88} & \textbf{45.49} & \textbf{85.67} & \textbf{49.90} & \textbf{88.78} & \textbf{21.34} & \textbf{56.21} & \textbf{10.82}\\ 
                \hline \hline
                \multirow{6}{*}{DN-121} & DIM & 51.10 & 6.01 & 99.60* & 98.70* & 44.19 & 4.61	& 63.43 & 3.51 & 29.06 & 1.70 \\
                & \textbf{DA-DIM} & \textbf{86.67} & \textbf{23.15} & \textbf{100.00*} & \textbf{100.00*} & \textbf{81.76} & \textbf{18.94} & \textbf{88.08} & \textbf{9.12} & \textbf{64.83} & \textbf{7.31}\\ 
                & TIM & 54.01 & 6.51 & 99.80* & 99.00* & 49.00 & 5.41 & 67.54 & 2.51 & 32.36	& 1.60\\ 
                & \textbf{DA-TIM} & \textbf{87.27} & \textbf{23.35} & \textbf{100.00*} & \textbf{100.00*} & \textbf{84.47} & \textbf{22.55} & \textbf{89.48} & \textbf{9.72} & \textbf{67.43} & \textbf{8.82}\\ 
                & SIT & 73.85 & 31.06 & \textbf{100.00*} & \textbf{100.00*} & 64.23 & 23.45 & 76.75 & 14.23 & 41.58 & 7.31 \\ 
                & \textbf{DA-SIT} & \textbf{90.18} & \textbf{44.99} & \textbf{100.00*} & 99.90* & \textbf{84.17} & \textbf{39.28} & \textbf{90.68} & \textbf{22.44} & \textbf{63.43} & \textbf{13.53}\\ 
                \hline \hline
                \multirow{6}{*}{WRN50-2} & DIM & 58.12 & 7.41 & 56.11 & 6.41 & 95.79* & 85.67* & 61.22 & 2.61 & 30.66 & 1.50\\
                & \textbf{DA-DIM} & \textbf{88.88} & \textbf{30.66} & \textbf{86.87} & \textbf{24.55} & \textbf{100.00*} & \textbf{99.70*} & \textbf{83.37} & \textbf{8.82} & \textbf{62.02} & \textbf{7.41}\\ 
                & TIM & 60.82 & 7.62 & 56.61 & 7.01 & 97.80* & 89.58* & 65.33 & 3.01 & 36.57 & 1.70\\ 
                & \textbf{DA-TIM} & \textbf{88.98} & \textbf{32.57} & \textbf{87.68} & \textbf{28.96} & \textbf{100.00*} & \textbf{99.90*} & \textbf{84.97} & \textbf{9.32} & \textbf{66.63} & \textbf{9.72} \\ 
                & SIT & 79.76 & 36.37 & 76.35 & 24.25 & 99.90* & 96.79* & 73.55 & 9.72 & 43.99 & 5.01\\ 
                & \textbf{DA-SIT} & \textbf{92.38} & \textbf{51.10} & \textbf{88.68} & \textbf{37.07} & \textbf{100.00*} & \textbf{97.70*} & \textbf{86.77} & \textbf{15.93} & \textbf{57.31} & \textbf{10.22} \\ 
                \hline \hline
                \multirow{6}{*}{VGG-19} & DIM & 38.78 & 1.00 & 39.68 & 1.00 & 31.06 & 0.6 & 99.40* & 94.49* & 21.04 & 0.3\\  
                & \textbf{DA-DIM} & \textbf{65.43} & \textbf{4.91} & \textbf{66.33} & \textbf{6.91} & \textbf{51.70} & \textbf{3.61} & \textbf{100.00*} & \textbf{99.80*} & \textbf{41.18} & \textbf{2.51} \\ 
                & TIM & 46.79 & 1.40 & 45.79 & 1.50 & 37.78 & 0.70 & 99.20* & 96.69* & 27.96 & 0.20\\  
                & \textbf{DA-TIM} & \textbf{66.13} & \textbf{6.51} & \textbf{70.54} & \textbf{8.12} & \textbf{58.32} & \textbf{5.01} & \textbf{100.00*} & \textbf{99.80*} & \textbf{48.60} & \textbf{3.51}\\
                & SIT & 49.40 & 4.31 & 51.30 & 6.21 & 38.68 & 3.01 & 99.90* & 97.70* & 26.85 & 1.80\\ 
                & \textbf{DA-SIT} & \textbf{63.43} & \textbf{8.32} & \textbf{64.93} & \textbf{9.82} & \textbf{48.00} & \textbf{4.91} & \textbf{100.00*} & \textbf{98.40*} & \textbf{36.57} & \textbf{3.31}\\
                \bottomrule \bottomrule
                \end{tabular}
        \caption{Fooling and targeted success rates (\%) of the adversarial examples generated by DIM, TIM, SIT and the corresponding IDAA-enhanced methods (\ie, DA-DIM, DA-TIM, and DA-SIT). ``*'' indicates the white-box setting.}
        \label{tab:int-attack}
        \end{center}
        \vskip -0.2in
        \end{table*}
        
In this section, we conduct extensive experiments on ImageNet to evaluate the proposed adversarial attack, IDAA.
We also explore IDAA by a series of ablation experiments.

\subsection{Experimental Setting}
\textbf{Dataset.} 
We randomly sample about 1000 images of different categories from the ILSVRC 2012 validation set~\cite{russakovsky2015imagenet}, which can be correctly classified by the adopted models in this paper.
For a given benign image, we randomly select a target class to conduct adversarial attacks.

\textbf{Networks.}
In this paper, we consider five naturally trained networks, \ie, ResNet-50 (RN-50)~\cite{xie2017aggregated}, VGG-19~\cite{simonyan2015very}, Inception-v3 (Inc-v3)~\cite{szegedy2016rethinking}, DenseNet-121 (DN-121)~\cite{huang2017densely}, and Wide-ResNet-50-2 (WRN50-2)~\cite{zagoruyko2016wide}, and two adversarially trained models, including adv-Inception-v3 ($\text{Inc-v3}_{adv}$) and ens-adv-Inception-ResNet-v2 ($\text{IncRes-v2}_{adv}$)~\cite{alex2018advinc3}.

\textbf{Baselines.}
We compare the proposed attack with five state-of-the-art adversarial attacks, including MI~\cite{dong2018boosting}, DIM~\cite{xie2019improving}, TIM~\cite{Dong_2019_CVPR}, Po-TRIP~\cite{li2020towards}, and SIT~\cite{wang2023structure}. 

\textbf{Parameters.}
For the hyper-parameters, we set the maximum perturbation to be $\epsilon=0.07$ with pixel values in $[0, 1]$ and the number of iterations $T$ is 10.
For the compared baselines, the step size is $\alpha=\epsilon/T$.
As IDAA has a much wider search space in $tanh$ space compared to the baselines, we choose a big step size, \ie, $\alpha=1.0$. 
For DIM and TIM, the diversity probability is set to be 0.7.
TIM exploits a Gaussian kernel with the size $7\times7$.
The decay factor $\mu$ is 1 for the momentum update.
SIA splits an image into $3\times3$ blocks.
The number of transformed images is 10 in SIA and IDAA. 
The exponential decay coefficients $\beta_1$ and $\beta_2$ in IDAA are set to 0.99 and 0.999, respectively.
The deviation strength $\gamma$ of IDAA is 0.1.
The mixup region occupies 0.7 of the original image, and the mixup operation is continuously executed three times. 

\textbf{Metrics.}
In this paper, we apply two metrics to evaluate the transferability.
\textit{1) fooling success rate} (fSuc), the percentage of adversarial examples that are misclassified by the victim model $f_b$.
\textit{2) targeted success rate} (tSuc), the percentage of adversarial examples that are classified as the target label by $f_b$.

\subsection{Attacking Naturally Trained Models}
    We first evaluate the attack performance on five naturally trained models, \ie, RN-50, DN-121, WRN50-2, VGG-19 and Inc-v3.
    The adversarial examples are crafted by RN-50, DN-121, WRN50-2, and VGG-19 with different attack methods, \ie, DIM, Po-TRIP, MI, TIM, SIT and IDAA.
    The generated adversarial examples are tested on the five models, respectively.
    \cref{tab:nat-attack} shows the attack performance of the generated adversarial attacks in the white-box and black-box settings.
    We can see that our attack method exhibits stronger transferability compared to the baselines.
    In the white-box setting, all the methods exhibit high attack performance. 
    Thus, we mainly focus on the transfer performance against the black-box models.
    For fooling success rate, our method outperforms DIM, Po-TRIP, MI, and TIM by a large margin of more than 20\%.
    In addition, the proposed method outperforms SIT by a clear margin of more than 10\% in terms of the fooling success rate.
    SIT has the best transferability among the baselines in terms of fooling success rate and targeted success rate.
    IDAA outperforms the best baseline by about 3\%$\sim$11\% in terms of the targeted success rate. 

    \begin{table*}[tbp]
        \begin{center}
        \small
                \begin{tabular}{lcccccccccccc}
                \toprule \toprule
                \multirow{4}{*}{Attack} & \multicolumn{10}{c}{Target Model} \\ 
                \cmidrule(lr){2-11}
                & \multicolumn{2}{c}{-RN-50} & \multicolumn{2}{c}{-DN-121} & \multicolumn{2}{c}{-WRN50-2} & \multicolumn{2}{c}{-VGG-19} & \multicolumn{2}{c}{-Inc-v3} & \multicolumn{2}{c}{Average}\\
                \cmidrule(lr){2-3}
                \cmidrule(lr){4-5}
                \cmidrule(lr){6-7}
                \cmidrule(lr){8-9}
                \cmidrule(lr){10-11}
                \cmidrule(lr){12-13}
                & fSuc & tSuc & fSuc & tSuc & fSuc & tSuc & fSuc & tSuc & fSuc & tSuc & fSuc & tSuc\\
                \midrule \midrule
                DIM & 73.15 & 22.95 & 73.85 & 23.35 & 67.43 & 20.94 & 77.35 & 11.82 & 47.39 & 9.22 & 67.83 & 16.83\\
                \hline
                Po-TRIP & 67.84 & 24.65 & 69.34 & 24.05 & 60.92 & 21.54 & 74.85 & 12.83 & 46.89 & 12.02 & 63.96 & 18.38\\
                \hline
                MI & 46.99	& 9.22 & 44.79 & 7.62 & 38.48 & 5.91 & 56.31 & 3.61 & 23.35 & 1.60 & 41.98 & 5.51\\
                \hline
                TIM & 59.12 & 14.73 & 56.81 & 14.23 & 50.60 & 13.53 & 67.33 & 5.71 & 35.07 & 3.31 & 53.78 & 9.49\\
                \hline
                SIT & 93.79 & 65.63 & 93.09 & 58.92 & 89.98 & 60.22 & 92.18 & 41.38 & 69.24 & 26.65 & 87.65 & 48.14\\
                \hline
                \textbf{IDAA} & \textbf{97.80} & \textbf{69.84} & \textbf{96.99} & \textbf{66.03} & \textbf{96.19} & \textbf{67.03} & \textbf{96.49} & \textbf{44.49} & \textbf{89.38} & \textbf{46.69} & \textbf{95.37} & \textbf{56.76}\\
                \bottomrule \bottomrule
                \end{tabular}
        \caption{Fooling and targeted success rates (\%) on five models in the ensemble setting. The adversarial examples are crafted on four of the ensemble models, \ie, RN-50, DN-121, WRN50-2, VGG-19 and Inc-v3. We hold out one of them as the black-box target model and craft adversarial examples by the remaining models, where ``-'' denotes the hold-out model.}
        \label{tab:ens-attack}
        \end{center}
        \vskip -0.2in
        \end{table*}

        \begin{table}[tbp]
            \begin{center}
                    \begin{tabular}{lcccc}
                    \toprule \toprule
                    \multirow{2}{*}{Attack} & \multicolumn{2}{c}{$\text{Inc-v3}_{adv}$} & \multicolumn{2}{c}{$\text{IncRes-v2}_{adv}$} \\
                    \cmidrule(lr){2-3}
                    \cmidrule(lr){4-5}
                    & fSuc & tSuc & fSuc & tSuc \\
                    \midrule \midrule
                    DIM & 29.06 & 2.10 & 17.74 & 0.70\\
                    \hline
                    Po-TRIP & 25.75 & 2.10 & 15.33 & 1.10\\
                    \hline
                    MI & 17.33 & 0.40 & 8.82 & 0.20\\
                    \hline
                    TIM & 24.05 & 0.80 & 15.13 & 0.50\\
                    \hline
                    SIT & 39.78 & 4.91 & 24.05 & 3.01\\
                    \hline
                    \textbf{IDAA} & \textbf{63.13} & \textbf{15.53} & \textbf{51.00} & \textbf{15.03}\\
                    \bottomrule \bottomrule
                    \end{tabular}
            \caption{Fooling and targeted success rates (\%) against the adversarially trained models. Adversarial examples are crafted by the ensemble models (\ie, RN-50, DN-121, WRN50-2, and VGG-19).}
            \label{tab:ad-attack}
            \end{center}
            \vskip -0.2in
            \end{table}

\subsection{Integrating into Other Input-Diversity-Based Attacks.}
    We integrate our methods into the input-diversity-based methods, \ie, DIM, TIM, and SIT, to evaluate that the proposed method could boost the transferability, which are denoted by DA-DIM, DA-TIM, and DA-ST, respectively.
    The fooling and targeted success rates are presented in \cref{tab:int-attack}.
    The results show that the attack performance in terms of the fooling success rate against black-box models is improved.
    In particular, the attacks integrated with our strategy improve the fooling success rate by about 10\%$\sim$30\%.
    The proposed strategy also boosts the targeted success rates of DIM, TIM, and SIT. 
    For instance, the targeted success rate of DA-SIT outperforms SIT with a large margin of 5\%$\sim$14\% when the surrogate model is WRN50-2. 
    The improvements demonstrate that our attack strategy (\ie, local mixup and adaptive step size) can help to boost the adversarial transferability.

\subsection{Attacking an Ensemble of Models.}
    As multiple white-box surrogate models can help to improve the transferability~\cite{dong2018boosting}, we exploit the ensemble attack to further improve the adversarial transferability.
    We generate the adversarial examples by averaging the logits of the ensemble models.
    Specifically, we conduct ensemble attacks on the five naturally trained models, \ie, RN-50, DN-121, WRN50-2, VGG-19, and Inc-v3.
    We hold out one of them as the black-box target model and average the logits of the other four models to craft the adversarial examples. 

    \cref{tab:ens-attack} shows the ensemble attacks' transfer results.
    We can see that the transferability of all the attacks is enhanced by ensemble attacks. 
    IDAA also keeps the best transfer performance compared to the baselines.
    In particular, armed with the ensemble attacks, IDAA can achieve an average fooling success rate of more than 95\% on different hold-out models. 
    The average targeted success rate of IDAA is about 56.76\%.
    IDAA outperforms the best baseline, SIT, by a large margin of more than 7\% in terms of the average fooling and targeted success rates.
    For other attacks, \ie, DIM, Po-TRIP, MI, and TIM, IDAA achieves 25\% higher average fooling and targeted success rates.

    \begin{figure*}[!t]
        \centering
        \begin{subfigure}{0.33\linewidth}
            \centerline{\includegraphics[width=\linewidth]{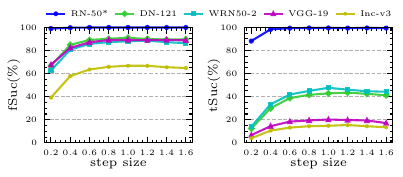}}
            \caption{RN-50}
            \label{subfig:lr-rn50}
        \end{subfigure}
        \begin{subfigure}{0.33\linewidth}
            \centerline{\includegraphics[width=\linewidth]{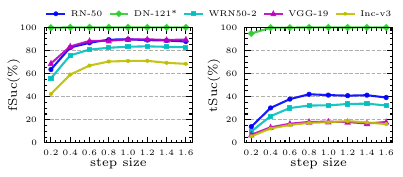}}
            \caption{DN-121}
            \label{subfig:lr-dn121}
        \end{subfigure}
        \begin{subfigure}{0.33\linewidth}
            \centerline{\includegraphics[width=\linewidth]{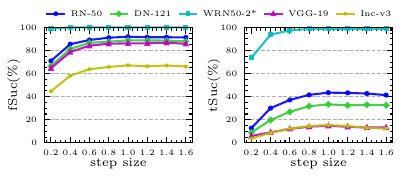}}
            \caption{WRN50-v2}
            \label{subfig:lr-wrn50}
        \end{subfigure}
        \caption{Fooling and targeted success rates (\%) against different step sizes. ``*'' is the white-box setting.}
        \label{fig:lr-attack}
        \vskip -0.1in
    \end{figure*}

    \begin{figure*}[!t]
        \centering
        \begin{subfigure}{0.33\linewidth}
            \centerline{\includegraphics[width=\linewidth]{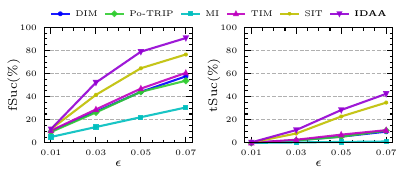}}
            \caption{RN-50 $\rightarrow$ DN-121}
            \label{subfig:eps-rn50}
        \end{subfigure}
        \begin{subfigure}{0.33\linewidth}
            \centerline{\includegraphics[width=\linewidth]{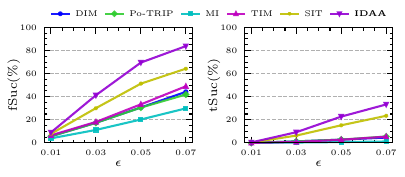}}
            \caption{DN-121 $\rightarrow$ WRN50-2}
            \label{subfig:eps-dn121}
        \end{subfigure}
        \begin{subfigure}{0.33\linewidth}
            \centerline{\includegraphics[width=\linewidth]{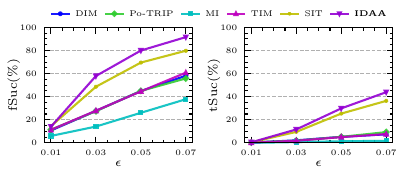}}
            \caption{WRN50-2 $\rightarrow$ RN-50}
            \label{subfig:eps-wrn50}
        \end{subfigure}
        \caption{Fooling and targeted success rates (\%) against different $\epsilon$ budgets.}
        \label{fig:eps-attack}
        \vskip -0.1in
    \end{figure*}

    \begin{figure*}[!t]
        \centering
        \begin{subfigure}{0.33\linewidth}
            \centerline{\includegraphics[width=\linewidth]{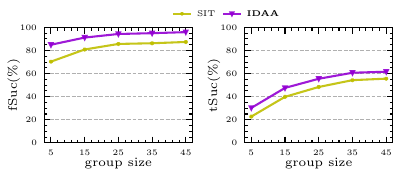}}
            \caption{RN-50 $\rightarrow$ DN-121}
            \label{subfig:eps-rn50}
        \end{subfigure}
        \begin{subfigure}{0.33\linewidth}
            \centerline{\includegraphics[width=\linewidth]{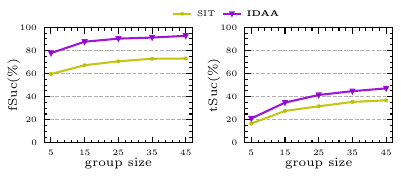}}
            \caption{DN-121 $\rightarrow$ WRN50-2}
            \label{subfig:eps-dn121}
        \end{subfigure}
        \begin{subfigure}{0.33\linewidth}
            \centerline{\includegraphics[width=\linewidth]{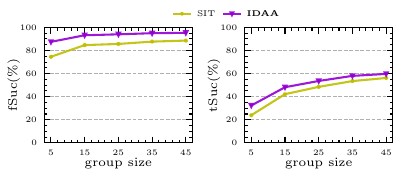}}
            \caption{WRN50-2 $\rightarrow$ RN-50}
            \label{subfig:eps-wrn50}
        \end{subfigure}
        \caption{Fooling and targeted success rates (\%) against different group sizes.}
        \label{fig:grp-attack}
        \vskip -0.1in
    \end{figure*}

\subsection{Attacking Adversarially Trained Models}
    Adversarial training is one of the defenses against adversarial attacks, which may largely mitigate the transferability of adversarial examples.
    In this subsection, we evaluate our proposed method against the adversarially trained models, \ie, $\text{Inc-v3}_{adv}$ and $\text{IncRes-v2}_{adv}$.
    Specifically, we first craft the adversarial examples by the ensemble models, including RN-50, DN-121, WRN50-2, and VGG-19.
    Then, we evaluate the fooling and targeted success rates against $\text{Inc-v3}_{adv}$ and $\text{IncRes-v2}_{adv}$.

    The attack performance against the adversarially trained models is shown in \cref{tab:ad-attack}.
    It can be observed that the transfer performance is reduced on the adversarially trained models.
    However, IDAA can still achieve a fooling success rate of more than 50\% on both models, while the fooling success rates of the baselines are about 8\%$\sim$40\%.
    The beast baseline, SIT, obtains the targeted success rate of about 3\%$\sim$5\%, and our method outperforms it by more than 10\%.

\subsection{Ablation Study}
In this subsection, we conduct ablation experiments to study the impacts of different settings, \ie, the step size, perturbation budget, and the group size of the transformed variants.
More ablation experiments are provided in the appendix, including the deviation strength $\gamma$, the number of mixups, the size of the mixup region, and the adaptive step size.

\textbf{Different step sizes.}
In this work, we project the perturbation into the $tanh$ space, which has a wider search space compared to the direct perturbation optimization.
Thus, IDAA needs a bigger step size to boost the search.
We evaluate the attack performance under the step sizes from 0.2 to 1.6.
The adversarial examples are generated by RN-50, DN-121, and WRN50-2, respectively.
The transfer performance is plotted in \cref{fig:lr-attack}.
We can see that a bigger step size (\eg, 1.0) helps to improve the attack performance in the white-box and black-box settings. 
A step size greater than 1.0 has no obvious benefits in enhancing the adversarial attacks, and a lower step size degrades the attack performance in terms of fooling and targeted success rates.

\textbf{Different $\epsilon$ budgets.}
The perturbation budget balances the tradeoff between the attack performance and the perceptible probability.
More perturbation budget means higher attack performance, but easier to be perceptible.
We evaluate the fooling and targeted success rates under the perturbation budgets 0.01, 0.03, 0.05, and 0.07.
As shown in \cref{fig:eps-attack}, the attack performance improves with the increase of the perturbation budgets. 
Besides, we observe that IDAA can always achieve better transferability under different budgets in comparison with the baselines.

\textbf{Different group sizes.}
Both SIA and IDAA exploit various image transformations to transform the adversarial example and obtain a group of transformed variants.
The group size decides the number of the transformed variants in one iterative step.
We evaluate the attack performance against different group sizes and the results are shown in ~\cref{fig:grp-attack}.
The fooling and targeted success rates boost with the increase in the group size.
When the group size is 25, it is hard to exploit a bigger group size to increase the fooling success rate. 
However, a bigger group size (\eg, 35) can further improve the targeted success rate even though the fooling success rate has no obvious change.
\section{Conclusion}
In this work, we propose a novel adversarial attack method, IDAA, combining a local mixup method and adaptive step size to strengthen the adversarial transferability. 
Specifically, we adopt various image transformations to transform an image to obtain the initial input diversity.
Local mixup operation is designed to mix the random regions among the transformed variants to further boost the diversity.
For precise perturbation optimization, we project the perturbation into the $tanh$ space and integrate the second-order momentum to adjust the step size dynamically.
Extensive experiments demonstrate that IDAA can achieve higher adversarial transferability in comparison with the competitive input-diversity-based methods.
Moreover, the proposed method can be integrated into other input-diversity-based attacks to further boost the transferability.
We hope that our proposed IDAA framework could shed light on crafting transferable adversarial examples in the black-box setting.
{
    \small
    \bibliographystyle{ieeenat_fullname}
    \bibliography{main}
}
\clearpage
\setcounter{page}{1}
\section*{Appendix}

\begin{table}[tbp]
            \begin{tabular}{ccccccc}
            \toprule \toprule
            \multirow{2}{*}{$\gamma$} & \multicolumn{2}{c}{\makecell{RN-50\\$\downarrow$\\DN-121}} & \multicolumn{2}{c}{\makecell{DN-121\\$\downarrow$\\WRN50-2}} & \multicolumn{2}{c}{\makecell{WRN50-2\\$\downarrow$\\RN-50}}\\
            \cmidrule(lr){2-3}
            \cmidrule(lr){4-5}
            \cmidrule(lr){6-7}
            & fSuc & tSuc & fSuc & tSuc & fSuc & tSuc\\
            \midrule \midrule
            0.00 & 78.66 & 36.57 & 67.13 & 25.05 & 82.46 & 38.08\\
            \hline
            0.05 & 86.47 & 42.48 & 80.16 & 32.16 & 88.48 & 43.19\\ 
            \hline
            0.10 & 90.18 & 44.69 & 83.87 & 32.87 & 90.98 & 43.79\\ 
            \hline
            0.15 & 91.28 & 43.79 & 86.97 & 32.97 & 92.38 & 42.38\\ 
            \hline
            0.20 & 93.19 & 43.99 & 89.38 & 32.16 & 93.19 & 42.59\\ 
            \hline
            0.25 & 94.29 & 41.88 & 89.48 & 32.16 & 94.29 & 41.98\\ 
            \bottomrule \bottomrule
            \end{tabular}
    \caption{Fooling and targeted success rates (\%) against different deviation strengths. We generate adversarial examples by RN-50, DN-121, and WRN50-2, respectively. The generated adversarial examples are tested on models DN-121, WRN50-2, and RN-50, respectively.}
    \label{tab:deviation-attack}
    \end{table}

    \begin{table}[tbp]
                \begin{tabular}{ccccccc}
                \toprule \toprule
                \multirow{2}{*}{Ratio} & \multicolumn{2}{c}{\makecell{RN-50\\$\downarrow$\\DN-121}} & \multicolumn{2}{c}{\makecell{DN-121\\$\downarrow$\\WRN50-2}} & \multicolumn{2}{c}{\makecell{WRN50-2\\$\downarrow$\\RN-50}}\\
                \cmidrule(lr){2-3}
                \cmidrule(lr){4-5}
                \cmidrule(lr){6-7}
                & fSuc & tSuc & fSuc & tSuc & fSuc & tSuc\\
                \midrule \midrule
                0.1 & 89.08 & 41.48 & 83.07 & 29.56 & 90.48 & 40.48\\
                \hline
                0.3	& 90.38 & 42.79 & 82.77 & 29.96 & 91.28 & 40.68\\
                \hline
                0.5	& 90.18 & 42.99 & 84.57 & 30.66 & 90.38 & 43.59\\
                \hline
                0.7	& 90.38 & 43.29 & 84.97 & 31.96 & 90.88 & 43.29\\
                \hline
                0.9	& 90.08 & 43.39 & 85.17 & 33.27 & 90.88 & 42.69\\
                \hline
                1.0	& 89.68 & 42.28 & 86.17 & 33.87 & 90.58 & 42.99\\
                \bottomrule \bottomrule
                \end{tabular}
        \caption{Fooling and targeted success rates (\%) against different sizes of mixup regions. We generate adversarial examples by RN-50, DN-121, and WRN50-2, respectively. The generated adversarial examples are tested on models DN-121, WRN50-2, and RN-50, respectively.}
        \label{tab:region-attack}
        \end{table}

        \begin{table}[tbp]
                    \begin{tabular}{ccccccc}
                    \toprule \toprule
                    \multirow{2}{*}{Time} & \multicolumn{2}{c}{\makecell{RN-50\\$\downarrow$\\DN-121}} & \multicolumn{2}{c}{\makecell{DN-121\\$\downarrow$\\WRN50-2}} & \multicolumn{2}{c}{\makecell{WRN50-2\\$\downarrow$\\RN-50}}\\
                    \cmidrule(lr){2-3}
                    \cmidrule(lr){4-5}
                    \cmidrule(lr){6-7}
                    & fSuc & tSuc & fSuc & tSuc & fSuc & tSuc\\
                    \midrule \midrule
                    0 & 89.38 & 39.28 & 82.67 & 28.86 & 87.37 & 29.46\\
                    \hline
                    1 & 90.38 & 43.29 & 84.97 & 31.96 & 90.88 & 43.29\\
                    \hline
                    2 & 90.28 & 43.79 & 84.57 & 33.47 & 91.38 & 44.49\\
                    \hline
                    3 & 90.18 & 44.79 & 85.47 & 34.17 & 91.28 & 42.99\\
                    \hline
                    4 & 89.58 & 42.28 & 83.77 & 33.67 & 90.08 & 41.98\\
                    \hline
                    5 & 87.98 & 39.08 & 82.67 & 32.57 & 90.38 & 39.28\\
                    \hline
                    6 & 86.77 & 37.88 & 79.76 & 30.36 & 90.18 & 38.28\\
                    \bottomrule \bottomrule
                    \end{tabular}
            \caption{Fooling and targeted success rates (\%) against different numbers of mixup. We generate adversarial examples by RN-50, DN-121, and WRN50-2, respectively. The generated adversarial examples are tested on models DN-121, WRN50-2, and RN-50, respectively. The sample region ratio occupies 0.7 of an image.}
            \label{tab:time-attack}
            \end{table}

\section{More Ablation Experiments}
\label{sec:more-ablation}
We conduct more ablation experiments to study the influence of the deviation strength $\gamma$, the size of the mixup region, the number of mixups, and the adaptive step size.

\subsection{Different Deviation Strengths}
In this work, we exploit a single triplet attack objective to conduct adversarial attacks.
Besides achieving the target goal, we also make the generated adversarial examples deviate from the ground-truth class and a coefficient $\gamma$ controls the deviation strength.
In this subsection, we evaluate IDAA against different deviation strengths.
The attack performance of IDAA is shown in \cref{tab:deviation-attack}.
We can see that the fooling success rate grows with the increase of $\gamma$, as adversarial examples deviate further from the ground-truth class.
A moderate deviation strength (\eg, $\gamma=0.10$) can help to improve the attack performance in terms of targeted success rate.
However, a too-big deviation strength (\eg, $\gamma=0.25$) will degrade the targeted success rate.
The reason may be that a too-big deviation strength disturbs the adversarial examples from getting close to the target class.  

\subsection{Different Sizes of the Mixup Regions}
The local mixup step randomly samples regions from a pair of transformed images to mix. 
The size of the sample region may influence the attack performance.
We evaluate IDAA against different sample ratios (\ie, 0.1, 0.3, 0.5, 0.7, 0.9, and 1.0) and execute a one-time mixup.
\cref{tab:region-attack} presents the attack performance of IDAA against different sizes of the mixup regions.
It can be observed that a big mixup region can further improve the transferability.
We speculate that a big mixup region can preserve important features of an image and increase the input diversity, enhancing the transferability.
In this work, we set the sample ratio to 0.7.

\begin{table*}[tbp]
    \begin{center}
    \small
            \begin{tabular}{lccccccccccc}
            \toprule \toprule
            \multirow{4}{*}{Surrogate} & \multirow{4}{*}{Attack} & \multicolumn{10}{c}{Target Model} \\
            \cmidrule(lr){3-12}
            & & \multicolumn{2}{c}{RN-50} & \multicolumn{2}{c}{DN-121} & \multicolumn{2}{c}{WRN50-2} & \multicolumn{2}{c}{VGG-19} & \multicolumn{2}{c}{Inc-v3} \\
            \cmidrule(lr){3-4}
            \cmidrule(lr){5-6}
            \cmidrule(lr){7-8}
            \cmidrule(lr){9-10}
            \cmidrule(lr){11-12}
            & & fSuc & tSuc & fSuc & tSuc & fSuc & tSuc & fSuc & tSuc & fSuc & tSuc\\
            \midrule \midrule
            \multirow{2}{*}{RN-50} & IDAA-MI & \textbf{100.00*} & 98.70* & \textbf{90.18} & 38.98 & \textbf{88.08} & 42.79 & 87.78 & 18.14 & \textbf{70.04} & 13.33 \\
            & IDAA-Ada & 99.90* & \textbf{99.50*} & 89.88 & \textbf{45.19} & 87.68 & \textbf{45.99} & 87.98 & \textbf{20.74} & 65.13 & \textbf{14.53} \\
            \hline \hline
            \multirow{2}{*}{DN-121} & IDAA-MI & 88.68 & 37.27 & \textbf{100.00*} & 99.70* & 82.97 & 30.36 & 88.08 & 17.64 & 68.34 & 15.23\\ 
            & IDAA-Ada & \textbf{89.58}	& \textbf{41.88} & \textbf{100.00*} & \textbf{100.00*} & \textbf{84.57} & \textbf{31.86} & \textbf{89.08} & \textbf{17.94} & \textbf{70.34} & \textbf{18.34}\\ 
            \hline \hline
            \multirow{2}{*}{WRN50-2} & IDAA-MI & 88.28 & 35.07 & 86.17 & 23.95 & \textbf{100.00*} & 96.09* & 83.37 & 12.22 & 65.33 & 9.82\\ 
            & IDAA-Ada & \textbf{91.08}	& \textbf{42.99} & \textbf{88.28} & \textbf{33.47} & \textbf{100.00*} & \textbf{98.80*} & \textbf{85.67} & \textbf{13.83} & \textbf{67.94} & \textbf{14.23}\\ 
            \bottomrule \bottomrule
            \end{tabular}
    \caption{Fooling and targeted success rates (\%) against identical and adaptive step sizes. IDAA with identical and adaptive step sizes are denoted by IDAA-MI and IDAA-Ada, respectively. IDAA-MI exploits the optimization step as the baseline MI. ``*'' is the white-box setting.}
    \label{tab:step-attack}
    \end{center}
    \vskip -0.2in
    \end{table*}

\subsection{Different Numbers of Mixups}
The local mixup operation may be executed more than one time.
Thus, we evaluate the influence of different execution times on IDAA.
We can see that the local mixup (\ie, the number of mixups is greater than 0) helps to boost the transferability.
Moreover, multiple local mixups may further improve the transferability.
For example, when the adversarial examples are generated by DN-121 and the local mixup is executed three times, the targeted success rate on WRN50-2 is improved to 34.14\% from 28.86\%.
However, too many local mixups (\ie, the number of mixups is greater than 4) will degrade the transferability.
The reason may be that the import features of an image are distorted by the regions of other images.
Thus, in this paper, we execute the local mixup three times.

\subsection{Adaptive Step Size}
In this subsection, we demonstrate the effect of the adaptive step size.
We use the identical and adaptive step sizes to update the perturbation of IDAA, respectively.
For the identical step size, we use the optimization step as the baseline MI, which is denoted as IDAA-MI.
To distinguish from IDAA-MI, we refer to our IDAA integrated with a second-order momentum as IDAA-Ada. 
The evaluation results are shown in \cref{tab:step-attack}.
The results demonstrate that the adaptive step size can help to improve the transferability, especially the targeted transferability.
For instance, when the surrogate and target models are WRN50-2 and RN-50, respectively, IDAA-Ada outperforms IDAA-MI by a clear margin of more than 7\% in terms of targeted success rate.

\begin{table*}[tbp]
    \begin{center}
    \small
            \begin{tabular}{lccccccccccc}
            \toprule \toprule
            \multirow{4}{*}{Surrogate} & \multirow{4}{*}{($\beta_1$, $\beta_2$)} & \multicolumn{10}{c}{Target Model} \\
            \cmidrule(lr){3-12}
            & & \multicolumn{2}{c}{RN-50} & \multicolumn{2}{c}{DN-121} & \multicolumn{2}{c}{WRN50-2} & \multicolumn{2}{c}{VGG-19} & \multicolumn{2}{c}{Inc-v3} \\
            \cmidrule(lr){3-4}
            \cmidrule(lr){5-6}
            \cmidrule(lr){7-8}
            \cmidrule(lr){9-10}
            \cmidrule(lr){11-12}
            & & fSuc & tSuc & fSuc & tSuc & fSuc & tSuc & fSuc & tSuc & fSuc & tSuc\\
            \midrule \midrule
            \multirow{10}{*}{RN-50} &  (0.99, 0.999) & 99.90* & 99.40* & 88.98 & 44.29 & 87.17 & 45.49 & 88.68 & 20.94 & 65.33 & 15.83 \\
            & (0.8, 0.89)	& 99.90* & 99.60* & 89.58 & 45.79 & 87.88 & 45.39 & 89.08 & 20.74 & 65.43 & 15.13\\
            & (0.7, 0.79) & \textbf{100.00*} & 99.70* & 90.68 & 47.39 & 89.38 & 49.50 & 90.48 & 21.74 & 67.74 & 16.53\\
            & (0.6, 0.69) & \textbf{100.00*} & 99.70* & 91.28 & 49.60 & 89.68 & 49.80 & 90.08 & 23.15 & 68.64 & 17.74\\
            & (0.5, 0.59) & \textbf{100.00*} & 99.80* & 91.98 & 51.20 & 90.38 & 54.01 & 91.48 & 24.55 & 68.74 & 19.34\\ 
            & (0.4, 0.49) & \textbf{100.00*} & \textbf{100.00*} & 92.08 & 54.51 & 91.28 & 57.41 & 91.48	& 25.95 & 69.84 & 21.04\\
            & (0.3, 0.39) & \textbf{100.00*} & \textbf{100.00*} & 93.19 & 56.41 & 91.88 & 59.92 & 91.98 & 28.66 & 70.84 & 21.54\\
            & (0.2, 0.29) & \textbf{100.00*} & \textbf{100.00*} & 93.69 & 59.02	& 92.48	& 61.22	& \textbf{93.59} & 29.96 & 72.85 & 23.15\\
            & (0.1, 0.19) & \textbf{100.00*} & \textbf{100.00*} & 94.49 & 62.32 & 93.89 & 64.23 & 93.49 & 31.46 & 73.75 & 24.15\\
            & \textbf{(0.0, 0.1)} & \textbf{100.00*} & \textbf{100.00*} & \textbf{94.89} & \textbf{63.23} & \textbf{94.39} & \textbf{66.03} & \textbf{93.59} & \textbf{32.97} & \textbf{74.95} & \textbf{26.85}\\
            \hline \hline
            \multirow{10}{*}{DN-121} & (0.99, 0.999) & 89.28 & 40.38 & \textbf{100.00*} & \textbf{100.00*} & 84.77 & 32.57 & 88.98 & 17.84 & 69.44 & 17.74\\
            & (0.8, 0.89) & 89.68 & 42.28 & \textbf{100.00*} & \textbf{100.00*} & 84.57 & 33.67	& 89.38	& 18.14	& 69.14	& 17.64\\
            & (0.7, 0.79) & 91.08 & 43.79 & \textbf{100.00*} & \textbf{100.00*} & 86.07 & 35.87	& 90.18	& 19.44	& 71.34	& 17.54\\
            & (0.6, 0.69) & 90.98 & 45.59 & \textbf{100.00*} & \textbf{100.00*} & 86.47 & 36.27	& 90.98	& 20.34	& 74.25	& 19.74\\
            & (0.5, 0.59) & 91.48 & 48.30 & \textbf{100.00*} & \textbf{100.00*} & 88.38 & 38.58 & 90.88	& 21.94	& 74.85	& 21.84\\
            & (0.4, 0.49) & 93.09 & 51.40 & \textbf{100.00*} & \textbf{100.00*} & 88.88 & 41.78 & 91.38	& 22.24	& 75.95	& 23.25\\
            & (0.3, 0.39) & 93.09 & 53.71 & \textbf{100.00*} & \textbf{100.00*} & 89.38 & 44.59	& 92.59	& 25.25	& 77.76	& 26.05\\
            & (0.2, 0.29) & 94.49 & 56.81 & \textbf{100.00*} & \textbf{100.00*} & 90.08 & 46.89 & 92.59 & 28.36 & 78.36 & 25.55\\
            & (0.1, 0.19) & 94.39 & 58.82 & \textbf{100.00*} & \textbf{100.00*} & 91.28 & 49.00	& 92.99 & \textbf{29.76} & \textbf{79.06} & 27.66\\
            & \textbf{(0.0, 0.1)} & \textbf{94.59} & \textbf{61.72} & \textbf{100.00*} & \textbf{100.00*} & \textbf{92.48} & \textbf{51.90}	& \textbf{93.49} & 29.66 & 78.76 & \textbf{29.16}\\ 
            \hline \hline
            \multirow{10}{*}{WRN50-2} & (0.99, 0.999) & 91.58 & 43.89 & 88.78 & 33.67 & \textbf{100.00*} & 98.70* & 86.87 & 14.03 & 67.84 & 14.53\\
            & (0.8, 0.89) & 91.88	& 45.79	& 89.58 & 36.97 & \textbf{100.00*} & 98.90* & 87.88	& 15.13	& 69.04 & 16.23\\
            & (0.7, 0.79) & 92.69	& 47.29	& 90.58 & 37.27 & \textbf{100.00*} & 99.60* & 87.07	& 15.93	& 69.74 & 17.43\\
            & (0.6, 0.69) & 93.19	& 50.30	& 90.48 & 41.58 & \textbf{100.00*} & 99.50* & 87.37	& 17.13	& 70.94 & 18.24\\
            & (0.5, 0.59) & 94.19	& 53.91	& 91.28 & 43.39 & \textbf{100.00*} & 99.90* & 87.98	& 17.33	& 72.34 & 18.74\\
            & (0.4, 0.49) & 94.29	& 57.82	& 92.18 & 47.09 & \textbf{100.00*} & 99.90* & 90.88	& 19.74	& 75.15 & 20.54\\
            & (0.3, 0.39) & 95.19	& 60.32	& 92.28 & 48.70 & \textbf{100.00*} & \textbf{100.00*} & 90.88	& 22.75	& 74.65 & 23.05\\
            & (0.2, 0.29) & 95.69	& 65.13	& 94.19 & 52.51 & \textbf{100.00*} & \textbf{100.00*} & 91.48	& 23.55	& 75.75 & 23.45\\
            & (0.1, 0.19) & \textbf{96.19}	& 67.33	& 94.49 & 56.31 & \textbf{100.00*} & \textbf{100.00*} & 90.78	& 24.25	& 77.56 & 25.85\\
            & \textbf{(0.0, 0.1)} & 95.79 & \textbf{69.34} & \textbf{95.09} & \textbf{56.61} & \textbf{100.00*} & \textbf{100.00*} & \textbf{91.78} & \textbf{27.66} & \textbf{78.46} & \textbf{28.26}\\ 
            \bottomrule \bottomrule
            \end{tabular}
    \caption{Fooling and targeted success rates (\%) against different $\beta_1$s and $\beta_2$s. ``*'' is the white-box setting.}
    \label{tab:beta12-attack}
    \end{center}
    \vskip -0.2in
    \end{table*}

    \begin{table*}[tbp]
        \begin{center}
        \small
                \begin{tabular}{lccccccccccc}
                \toprule \toprule
                \multirow{4}{*}{Surrogate} & \multirow{4}{*}{($\beta_1$, $\beta_2$)} & \multicolumn{10}{c}{Target Model} \\
                \cmidrule(lr){3-12}
                & & \multicolumn{2}{c}{RN-50} & \multicolumn{2}{c}{DN-121} & \multicolumn{2}{c}{WRN50-2} & \multicolumn{2}{c}{VGG-19} & \multicolumn{2}{c}{Inc-v3} \\
                \cmidrule(lr){3-4}
                \cmidrule(lr){5-6}
                \cmidrule(lr){7-8}
                \cmidrule(lr){9-10}
                \cmidrule(lr){11-12}
                & & fSuc & tSuc & fSuc & tSuc & fSuc & tSuc & fSuc & tSuc & fSuc & tSuc\\
                \midrule \midrule
                \multirow{10}{*}{RN-50} & (0.0, 0.9) & 99.90* & 99.60* & 86.67 & 35.47 & 85.77 & 38.18 & 87.68 & 16.83 & 61.42 & 11.52\\
                & (0.0, 0.8) & \textbf{100.00*} & 99.90* & 91.08 & 47.09 & 89.68 & 47.90 & 90.18 & 22.44 & 66.63 & 16.53\\
                & (0.0, 0.7) & \textbf{100.00*} & 99.90* & 92.59 & 54.71 & 90.88 & 56.31 & 92.08 & 26.35 & 71.54 & 19.94\\
                & (0.0, 0.6) & \textbf{100.00*} & \textbf{100.00*} & 93.29 & 57.62 & 92.99 & 59.32 & 91.98 & 29.46 & 71.74 & 22.55\\
                & (0.0, 0.5) & \textbf{100.00*} & \textbf{100.00*} & 94.59 & 61.92 & 92.99 & 61.92 & 92.89 & 29.36 & 72.04 & 23.15\\
                & (0.0, 0.4) & \textbf{100.00*} & \textbf{100.00*} & 94.39 & 61.62 & 93.19 & 63.83 & 92.08 & 31.06 & 72.85 & 24.05\\
                & (0.0, 0.3) & \textbf{100.00*} & \textbf{100.00*} & \textbf{95.29} & 62.42 & 94.09 & 64.63 & 93.29 & 32.36 & 72.75 & 25.85\\
                & (0.0, 0.2) & \textbf{100.00*} & \textbf{100.00*} & 94.69 & 62.63 & 94.19 & 65.83 & \textbf{93.79} & 31.86 & 74.15 & 25.15 \\ 
                & \textbf{(0.0, 0.1)} & \textbf{100.00*} & \textbf{100.00*} & 94.89 & \textbf{63.23} & \textbf{94.39} & \textbf{66.03} & 93.59 & \textbf{32.97} & \textbf{74.95} & \textbf{26.85}\\
                & (0.0, 0.0) & \textbf{100.00*} & \textbf{100.00*} & 94.19 & 62.42 & 92.89 & 62.73 & 92.38 & 30.66 & 70.94 & 23.95\\
                \hline \hline
                \multirow{10}{*}{DN-121} & (0.0, 0.9) & 86.77 & 34.87 & \textbf{100.00*} & \textbf{100.00*} & 82.97 & 27.76 & 88.88 & 14.53 & 66.13 & 14.83\\
                & (0.0, 0.8) & 91.78 & 45.19 & \textbf{100.00*} & \textbf{100.00*} & 86.77 & 36.67 & 91.28 & 19.84 & 71.44 & 19.14\\
                & (0.0, 0.7) & 92.69 & 51.50 & \textbf{100.00*} & \textbf{100.00*} & 88.98 & 42.48 & 91.18 & 23.55 & 75.75 & 22.65\\
                & (0.0, 0.6) & 93.59 & 55.61 & \textbf{100.00*} & \textbf{100.00*} & 90.48 & 44.49 & 92.18 & 25.55 & 76.55 & 23.75\\
                & (0.0, 0.5) & 94.49 & 57.82 & \textbf{100.00*} & \textbf{100.00*} & 91.48 & 47.19 & 93.49 & 26.65 & 77.56 & 25.85\\
                & (0.0, 0.4) & 94.69 & 59.32 & \textbf{100.00*} & \textbf{100.00*} & 91.88 & 50.20 & 92.89 & 27.35 & 77.86 & 26.85\\
                & (0.0, 0.3) & \textbf{95.19} & 59.62 & \textbf{100.00*} & \textbf{100.00*} & 92.38 & 49.70 & 92.48 & 28.26 & \textbf{79.96} & 26.35\\
                & (0.0, 0.2) & 94.89 & 60.82 & \textbf{100.00*} & \textbf{100.00*} & 92.08 & 50.50 & 93.39 & 28.56 & 79.06 & 28.56\\ 
                & \textbf{(0.0, 0.1)} & 94.59 & \textbf{61.72} & \textbf{100.00*} & \textbf{100.00*} & \textbf{92.48} & \textbf{51.90} & \textbf{93.49} & \textbf{29.66} & 78.76 & \textbf{29.16}\\ 
                & (0.0, 0.0) & 93.79 & 60.52 & \textbf{100.00*} & \textbf{100.00*} & 90.98 & 49.60 & 92.79 & 26.05 & 76.55 & 26.85\\
                \hline \hline
                \multirow{10}{*}{WRN50-2} & (0.0, 0.9) & 90.58 & 40.58 & 88.18 & 32.06 & \textbf{100.00*} & 99.10* & 85.97 & 13.33 & 66.83 & 12.63\\
                & (0.0, 0.8) & 94.19 & 53.01 & 91.28 & 42.28 & \textbf{100.00*} & 99.80* & 88.78 & 17.43 & 71.54 & 18.54\\
                & (0.0, 0.7) & 94.69 & 60.52 & 93.19 & 48.50 & \textbf{100.00*} & 99.90* & 90.08 & 20.74 & 75.85 & 23.05\\
                & (0.0, 0.6) & 95.49 & 63.13 & 93.79 & 52.71 & \textbf{100.00*} & 99.90* & 91.08 & 23.35 & 76.55 & 22.95\\
                & (0.0, 0.5) & 95.99 & 64.43 & 94.69 & 53.91 & \textbf{100.00*} & \textbf{100.00*} & 91.78 & 25.55 & 78.26 & 26.15\\
                & (0.0, 0.4) & 96.09 & 66.83 & \textbf{95.09} & 55.81 & \textbf{100.00*} & \textbf{100.00*} & 91.98 & 26.95 & 77.66 & 27.56\\
                & (0.0, 0.3) & \textbf{96.59} & 68.34 & 94.59 & 57.21 & \textbf{100.00*} & \textbf{100.00*} & \textbf{92.28} & 27.15 & \textbf{78.46} & 27.15\\
                & (0.0, 0.2) & 96.19 & 69.04 & 94.19 & \textbf{58.02} & \textbf{100.00*} & \textbf{100.00*} & \textbf{92.28} & 26.65 & 77.45 & 27.05\\ 
                & \textbf{(0.0, 0.1)} & 95.79 & \textbf{69.34} & \textbf{95.09} & 56.61 & \textbf{100.00*} & \textbf{100.00*} & 91.78 & \textbf{27.66} & \textbf{78.46} & \textbf{28.26}\\ 
                & (0.0, 0.0) & 95.69 & 68.24 & 94.49 & 57.41 & \textbf{100.00*} & \textbf{100.00*} & 91.18 & 25.75 & 75.85 & 28.16\\
                \bottomrule \bottomrule
                \end{tabular}
        \caption{Fooling and targeted success rates (\%) against different $\beta_2$s ($\beta_1$ is set to 0.0). ``*'' is the white-box setting.}
        \label{tab:beta2-attack}
        \end{center}
        \vskip -0.2in
        \end{table*}

\subsection{Different $\beta_1$s and $\beta_2$s}
In the update step, two hyper-parameters $\beta_1$ and $\beta_2$ decide the decay rates.
We conduct experiments to evaluate their influences on the generated adversarial examples.
We first evaluate the transferability performance under different $\beta_1$s and $\beta_2$s, as shown in \cref{tab:beta12-attack}.
We can observe that the transferability performance grows with the decrease of $\beta_1$ and $\beta_2$.
When $\beta_1$ and $\beta_2$ are set to 0.0 and 0.1, respectively, the attack performance of IDAA can be improved by a large margin in terms of the fooling and targeted success rates.
For instance, the adversarial examples generated by RN-50 achieve a fooling success rate of 74.95\% and a targeted success rate of 26.85\% on Inc-v3, respectively. 
When contrasted with adversarial examples generated using $\beta_1=0.99$ and $\beta_2=0.999$, the fooling and targeted success rates exhibit improvements exceeding 9\% and 11\%, correspondingly.

To further assess the impact of the hyperparameter $\beta_2$, we set $\beta_1$ to 0.0 and vary $\beta_2$ across the range from 0.0 to 0.9 in increments of 0.1.
The experimental results are shown in \cref{tab:beta2-attack}.
From \cref{tab:beta2-attack}, we know that a small $\beta_2$ can benefit on the improvement of attack transferability.
However, the attack performance will degrade when $\beta_2$ is set to 0.0 in comparison to the adversarial examples generated by $\beta_2=0.1$.
Thus, for better attack performance, we can choose to set the $\beta_1$ and $\beta_2$ to 0.0 and 0.1, respectively.


\end{document}